\documentclass[english]{article}
\usepackage{spconf}
\usepackage{amsmath}
\usepackage{url}

\title{Advances in Optimizing Recurrent Networks}

\name{Yoshua Bengio, Nicolas Boulanger-Lewandowski and Razvan Pascanu}
\address{U. Montreal}

\begin{document}
\ninept
\maketitle
\begin{abstract}
After a more than decade-long period of relatively little research
activity in the area of recurrent neural networks, several new developments
will be reviewed here that have allowed substantial progress both in
understanding and in technical solutions towards more efficient training of
recurrent networks.  These advances have been motivated by and related to
the optimization issues surrounding deep learning. Although recurrent
networks are extremely powerful in what they can in principle represent in
terms of modeling sequences, their training is plagued by two aspects of
the same issue regarding the learning of long-term dependencies. Experiments
reported here evaluate the use of clipping gradients, spanning longer time
ranges with leaky integration, advanced momentum techniques, using more
powerful output probability models, and encouraging sparser gradients to
help symmetry breaking and credit assignment. The experiments are performed
on text and music data and show off the combined effects of these
techniques in generally improving both training and test error.
\end{abstract}


\section{Introduction}

Machine learning algorithms for capturing statistical structure in
sequential data face a fundamental
problem~\cite{Hochreiter91-small,Bengio-trnn93-small}, called the {\em difficulty of
  learning long-term dependencies}. If the operations performed when
forming a fixed-size summary of relevant past observations (for the purpose
of predicting some future observations) are linear, this summary must
exponentially forget past events that are further away, to maintain
stability. On the other hand, if they are non-linear, then this
non-linearity is composed many times, yielding a highly non-linear
relationship between past events and future events.  Learning such
non-linear relationships turns out to be difficult, for reasons that are
discussed here, along with recent proposals for reducing this difficulty.

Recurrent neural networks~\cite{Rumelhart86b-small} can represent such non-linear
maps ($F$, below) that iteratively build a relevant summary of past
observations. In their simplest form, recurrent neural networks (RNNs) form a
deterministic {\em state} variable $h_t$ as a function of the present input
observation $x_t$ and the past value(s) of the state variable, e.g., $h_t =
F_\theta(h_{t-1},x_t)$, where $\theta$ are tunable parameters that control
what will be remembered about the past sequence and what will be discarded.
Depending on the type of problem at hand, a loss function $L(h_t,y_t)$ is
defined, with $y_t$ an observed random variable at time $t$ and
$C_t=L(h_t,y_t)$ the cost at time $t$. The generalization objective is to
minimize the expected future cost, and the training objective involves the
average of $C_t$ over observed sequences.  In principle, RNNs can
be trained by gradient-based optimization procedures (using the
back-propagation algorithm~\cite{Rumelhart86b-small} to compute a gradient), but
it was observed early on~\cite{Hochreiter91-small,Bengio-trnn93-small} that capturing
dependencies that span a long interval was difficult, making the task of
optimizing $\theta$ to minimize the average of $C_t$'s almost impossible
for some tasks when the span of the dependencies of interest increases
sufficiently. More precisely, using a local numerical optimization such as
stochastic gradient descent or second order methods (which gradually
improve the solution), the proportion of trials (differing only from their
random initialization) falling into the basin of attraction of a good
enough solution quickly becomes very small as the temporal span of
dependencies is increased (beyond tens or hundreds of steps, depending of
the task).

These difficulties are probably responsible for the major reduction in
research efforts in the area of RNNs in the 90's and 2000's.
However, a revival of interest in these learning algorithms is taking place,
in particular thanks to \cite{Martens+Sutskever-ICML2011-small} and \cite{Mikolov-ICASSP-2011-small}.
This paper studies the issues giving rise to these difficulties and
discusses, reviews, and combines several techniques that have been proposed 
in order to improve training of RNNs, following up on
a recent thesis devoted to the subject~\cite{Sutskever-thesis2012-small}.
We find that these techniques generally help generalization performance as well
as training performance, which suggest they help to improve the optimization
of the training criterion. We also find that although these techniques
can be applied in the online setting, i.e., as add-ons to stochastic
gradient descent (SGD), they allow to compete with batch (or large minibatch)
second-order methods such as Hessian-Free optimization, recently found
to greatly help training of RNNs~\cite{Martens+Sutskever-ICML2011-small}.

\section{Learning Long-Term Dependencies and the Optimization Difficulty with Deep Learning}

There has been several breakthroughs in recent years in the algorithms and
results obtained with so-called {\em deep learning} algorithms
(see~\cite{Bengio-2009-book} and \cite{Bengio-Courville-Vincent-arxiv-2012-small}
for reviews). Deep learning algorithms discover multiple levels of
representation, typically as deep neural networks or graphical models
organized with many levels of representation-carrying latent variables.
Very little work on deep architectures occurred before the major advances of
2006~\cite{Hinton06-small,Bengio-nips-2006-small,ranzato-07-small}, probably because of {\em
  optimization difficulties} due to the high level of non-linearity in
deeper networks (whose output is the composition of the non-linearity at
each layer). Some experiments~\cite{Erhan+al-2010-small} showed the
presence of an extremely large number of apparent local minima of the
training criterion, with no two different initializations going to the same
{\em function} (i.e. eliminating the effect of permutations and other
symmetries of parametrization giving rise to the same function).
Furthermore, qualitatively different initialization (e.g., using
unsupervised learning) could yield models in completely different regions
of function space. An unresolved question is whether these difficulties are
actually due to local minima or to ill-conditioning (which makes gradient
descent converge so slowly as to appear stuck in a local minimum).  Some
ill-conditioning has clearly been shown to be involved, especially for the
difficult problem of training deep auto-encoders, through
comparisons~\cite{martens2010hessian-small} of stochastic gradient descent and
Hessian-free optimization (a second order optimization method).
These optimization questions become particularly important when
trying to train very large networks on very large datasets~\cite{QuocLe-ICML2012-small},
where one realizes that a major challenge for deep learning is the
{\em underfitting} issue. Of course one can trivially overfit by increasing
capacity in the wrong places (e.g. in the output layer), but what we are
trying to achieve is learning of more powerful representations in order
to also get good generalization.

The same questions can be asked for RNNs. When the
computations performed by a RNN are unfolded through
time, one clearly sees a deep neural network {\em with shared weights}
(across the 'layers', each corresponding to a different time step), and
with a cost function that may depends on the output of intermediate layers.
Hessian-free optimization has been successfully used to considerably
extend the span of temporal dependencies that a RNN
can learn~\cite{Martens+Sutskever-ICML2011-small}, suggesting that ill-conditioning
effects are also at play in the difficulties of training RNN.

An important aspect of these difficulties is that the gradient can be
decomposed~\cite{Bengio-trnn93-small,Pascanu-arxiv-2012} into terms that involve
{\em products of Jacobians} $\frac{\partial h_t}{\partial h_{t-1}}$ over
subsequences linking an event at time $t_1$ and one at time $t_2$:
$\frac{\partial h_{t_2}}{\partial h_{t_1}}=\prod_{\tau=t_1+1}^{t_2} \frac{\partial h_{\tau}}{\partial h_{\tau-1}}$.
As $t_2-t_1$ increases, the products of $t_2-t_1$ of these Jacobian matrices
tend to either vanish (when the leading eigenvalues of $\frac{\partial
  h_t}{\partial h_{t-1}}$ are less than 1) or explode (when the leading
eigenvalues of $\frac{\partial h_t}{\partial h_{t-1}}$ are greater than 1\footnote{
Note that this is not a sufficient condition, but a necessary one. Further more one usually 
wants to operate in the regime where the leading eigenvalue is larger than 1 but the gradients do not explode.}).
This is problematic because the total gradient due to a loss $C_{t_2}$ at time $t_2$
is a sum whose terms correspond to the effects at different time spans, which are weighted 
by $\frac{\partial h_{t_2}}{\partial h_{t_1}}$ for different $t_1$'s:
\[
  \frac{\partial C_{t_2}}{\partial \theta} = \sum_{t_1\leq t_2} \frac{\partial C_{t_2}}{\partial h_{t_2}}
 \frac{\partial h_{t_2}}{\partial h_{t_1}}\frac{\partial h_{t_1}}{\partial \theta^{(t_1)}}
\]
where $\frac{\partial h_{t_1}}{\partial \theta^{(t_1)}}$ is the derivative
of $h_{t_1}$ with respect to the instantiation of the parameters $\theta$
at step $t_1$, i.e., that directly come into the computation of $h_{t_1}$
in $F$. When the $\frac{\partial h_{t_2}}{\partial h_{t_1}}$ tend to vanish
for increasing $t_2-t_1$, the long-term term effects become exponentially
smaller in magnitude than the shorter-term ones, making it very difficult
to capture them. On the other hand, when $\frac{\partial h_{t_2}}{\partial
  h_{t_1}}$ ``explode'' (becomes large), gradient descent updates can be
destructive (move to poor configuration of parameters). It is not that the
gradient is wrong, it is that gradient descent makes small but finite steps
$\Delta \theta$ yielding a $\Delta C$, whereas the gradient measures the
effect of $\Delta C$ when $\Delta \theta \rightarrow 0$.  A much deeper
discussion of this issue can be found in \cite{Pascanu-arxiv-2012}, along
with a point of view inspired by dynamical systems theory and by the
geometrical aspect of the problem, having to do with the shape of the
training criterion as a function of $\theta$ near those regions of
exploding gradient. In particular, it is argued that the strong
non-linearity occurring where gradients explode is shaped like a cliff where
not just the first but also the second derivative becomes large {\em in the
  direction orthogonal to the cliff}. Similarly, flatness of the cost
function occurs simultaneously on the first and second derivatives.  Hence
dividing the gradient by the second derivative in each direction
(i.e., pre-multiplying by the inverse of some proxy for the Hessian matrix)
could in principle reduce the exploding and vanishing gradient effects, as
argued in~\cite{Martens+Sutskever-ICML2011-small}.

\section{Advances in Training Recurrent Networks} \label{sec:advances}

\subsection{Clipped Gradient}

To address the exploding gradient effect, \cite{Mikolov-thesis-2012,Pascanu-arxiv-2012} recently
proposed to {\em clip gradients above a given threshold}. Under the hypothesis that the explosion
occurs in very small regions (the cliffs in cost function mentioned above), most of the time
this will have no effect, but it will avoid aberrant parameter changes in those cliff regions,
while guaranteeing that the resulting updates are still in a descent direction. The specific
form of clipping used here was proposed in \cite{Pascanu-arxiv-2012} and is discussed there
at much greater length: when the norm of the gradient vector $g$ for a given sequence is above a $threshold$,
the update is done in the direction $threshold \frac{g}{||g||}$. As argued in \cite{Pascanu-arxiv-2012},
this very simple method implements a very simple form of second order optimization in the sense that
the second derivative is also proportionally large in those exploding gradient regions.

\subsection{Spanning Longer Time Ranges with Leaky Integration}

An old idea to reduce the effect of vanishing gradients is to introduce
shorter paths between $t_1$ and $t_2$, either via connections with longer
time delays \cite{Lin95-small} or inertia (slow-changing units) in some of the
hidden units~\cite{Elhihi+Bengio-nips8-small, JaegerLPS07}, or
both~\cite{Sutskever+Hinton-2009-small}.  Long-Short-Term Memory (LSTM)
networks~\cite{Hochreiter+Schmidhuber-1997}, which were shown to be able to
handle much longer range dependencies, also benefit from a linearly
self-connected memory unit with a near 1 self-weight which allows signals
(and gradients) to propagate over long time spans.

A different interpretation to this slow-changing units is that they behave like 
low-pass filter and hence they can be used to focus certain units on different 
frequency regions of the data. The analogy can be 
brought one step further by introducing band-pass filter units \cite{Siewert2007}
or by using domain specific knowledge to decide on what frequency bands
different units should focus. \cite{Mikolov+Zweig-2012} shows that adding low frequency information 
as an additional input to a recurrent network helps improving the performance of the model. 

In the experiments
performed here, a subset of the units were forced to change slowly
by using the following ``leaky integration'' state-to-state map:
$h_{t,i} = \alpha_i h_{t-1,i} + (1-\alpha_i) F_i(h_{t-1},x_t)$. The standard
RNN corresponds to $\alpha_i=0$, while here different values
of $\alpha_i$ were randomly sampled from $(0.02,0.2)$, allowing some units
to react quickly while others are forced to change slowly, but also
propagate signals and gradients further in time. Note that because
$\alpha < 1$, the vanishing effect is still present (and gradients
can still explode via $F$), but the {\em time-scale} of the vanishing
effect can be expanded.

\subsection{Combining Recurrent Nets with a Powerful Output Probability Model}

One way to reduce the underfitting of RNNs is to introduce
multiplicative interactions in the parametrization of $F$, as was done
successfully in~\cite{Martens+Sutskever-ICML2011-small}. When the output predictions
are multivariate, another approach
is to capture the high-order dependencies between the output variables
using a powerful output probability model such as a Restricted
Boltzmann Machine (RBM)~\cite{SutskeverHintonTaylor2009-small,Boulanger+al-ICML2012-small}
or a deterministic variant of it called NADE~\cite{Larochelle+Murray-2011-small,Boulanger+al-ICML2012-small}.
In the experiments performed here, we have experimented with a NADE output model
for the music data.

\subsection{Sparser Gradients via Sparse Output Regularization and Rectified Outputs}

\cite{Bengio-2009-book} hypothesized that one reason for the difficulty in optimizing
deep networks is that in ordinary neural networks gradients diffuse through the layers,
diffusing credit and blame through many units, maybe making it difficult for hidden
units to specialize. When the gradient on hidden units is more sparse, one could
imagine that symmetries would be broken more easily and credit or blame assigned
less uniformly. This is what was advocated in \cite{Glorot+al-AI-2011-small}, 
exploiting the idea of rectifier non-linearities introduced earlier in \cite{Nair-2010-small},
i.e., the neuron non-linearity is $out = \max(0, in)$ instead of $out = \tanh(in)$ or
$out = {\rm sigmoid}(in)$.
This approach was very successful in recent work on deep learning for object
recognition~\cite{Krizhevsky-2012-small}, beating by far the state-of-the-art on ImageNet
(1000 classes). Here, we apply this deep learning idea to RNNs,
using an L1 penalty on outputs of hidden units to promote sparsity of activations.
The underlying hypothesis is that if the gradient is concentrated in a few paths
(in the unfolded computation graph of the RNN), it will reduce
the vanishing gradients effect.

\subsection{Simplified Nesterov Momentum}

Nesterov accelerated gradient (NAG)~\cite{Nesterov83} is a first-order optimization method to improve stability and convergence of regular gradient descent.
Recently, \cite{Sutskever-thesis2012-small} showed that NAG could be computed by the following update rules:
\begin{align}
v_t &= \mu_{t-1} v_{t-1} - \epsilon_{t-1} \nabla f(\theta_{t-1} + \mu_{t-1}v_{t-1}) \label{eq:sutskever_up_v} \ \\
\theta_t &= \theta_{t-1} + v_t \label{eq:sutskever_up_t}
\end{align}
where $\theta_t$ are the model parameters, $v_t$ the velocity, $\mu_t\in[0,1]$ the momentum (decay) coefficient and $\epsilon_t>0$ the learning rate at iteration $t$, $f(\theta)$ is the objective function and $\nabla f(\theta')$ is a shorthand notation for the gradient $\frac{\partial f(\theta)}{\partial\theta}|_{\theta=\theta'}$.
These equations have a form similar to standard momentum updates:
\begin{align}
v_t &= \mu_{t-1} v_{t-1} - \epsilon_{t-1} \nabla f(\theta_{t-1}) \label{eq:momentum_up_v} \\
\theta_t &= \theta_{t-1} + v_t  \label{eq:momentum_up_t} \\
&=  \theta_{t-1} + \mu_{t-1} v_{t-1} - \epsilon_{t-1} \nabla f(\theta_{t-1}) \label{eq:momentum_up_t2}
\end{align}
and differ only in the evaluation point of the gradient at each iteration.
This important difference, thought to counterbalance too high velocities by ``peeking ahead" actual objective values in the candidate search direction, results in significantly improved RNN performance on a number of tasks.

In this section, we derive a new formulation of Nesterov momentum differing from (\ref{eq:momentum_up_v}) and (\ref{eq:momentum_up_t2}) only in the linear combination coefficients of the velocity and gradient contributions at each iteration, and we offer an alternative interpretation of the method.
The key departure from (\ref{eq:sutskever_up_v}) and (\ref{eq:sutskever_up_t}) resides in committing to the ``peeked-ahead" parameters $\Theta_{t-1}\equiv\theta_{t-1} + \mu_{t-1}v_{t-1}$ and backtracking by the same amount before each update.
Our new parameters $\Theta_t$ updates become:
\begin{align}
v_t &= \mu_{t-1} v_{t-1} - \epsilon_{t-1} \nabla f(\Theta_{t-1}) \label{eq:new} \\
\Theta_t &= \Theta_{t-1} -\mu_{t-1}v_{t-1} + \mu_t v_t + v_t \nonumber \\
&= \Theta_{t-1} + \mu_t\mu_{t-1} v_{t-1} - (1+\mu_t)\epsilon_{t-1} \nabla f(\Theta_{t-1}) \label{eq:new_up_t}
\end{align}
Assuming a zero initial velocity $v_1=0$ and velocity at convergence of optimization $v_T\simeq 0$, the parameters $\Theta$ are a completely equivalent replacement of $\theta$.

Note that equation (\ref{eq:new_up_t}) is identical to \emph{regular} momentum~(\ref{eq:momentum_up_t2}) with different linear combination coefficients.
More precisely, for an equivalent velocity update (\ref{eq:new}), the velocity contribution to the new parameters $\mu_t\mu_{t-1} < \mu_t$ is reduced relatively to the gradient contribution $(1+\mu_t)\epsilon_{t-1}>\epsilon_{t-1}$.
This allows storing past velocities for a longer time with a higher $\mu$, while actually using those velocities more conservatively during the updates.
We suspect this mechanism is a crucial ingredient for good empirical performance.
While the ``peeking ahead" point of view suggests that a similar strategy could be adapted for regular gradient descent (misleadingly, because it would amount to a reduced learning rate $\epsilon_t$), our derivation shows why it is important to choose search directions aligned with the current velocity to yield substantial improvement.
The general case is also simpler to implement.

\section{Experiments}
\label{sec:datasets}

In the experimental section we compare vanilla SGD versus SGD plus some of
the enhancements discussed above. Specifically we use the letter `C` to
indicate that gradient clipping is used, `L` for leaky-integration units,
`R` if we use rectifier units with L1 penalty and `M` for Nesterov momentum.

\subsection{Music Data}
We evaluate our models on the four polyphonic music datasets of varying complexity used in \cite{Boulanger+al-ICML2012-small}: classical piano music (Piano-midi.de), folk 
tunes with chords instantiated from ABC notation (Nottingham), orchestral music (MuseData) and the four-part chorales by J.S.~Bach (JSB chorales).
The symbolic sequences contain high-level pitch and timing information in the form of a binary matrix, or \emph{piano-roll}, specifying precisely which notes occur at each time-step.
They form interesting benchmarks for RNNs because of their high dimensionality and the complex temporal dependencies involved at different time scales.
Each dataset contains at least 7 hours of polyphonic music with an average polyphony (number of simultaneous notes) of 3.9.

Piano-rolls were prepared by aligning each time-step (88 pitch labels that cover the whole range of piano) on an integer fraction of the beat (quarter note) 
and transposing each sequence in a common tonality (C major/minor) to facilitate learning.
Source files and preprocessed piano-rolls split in train, validation and test sets are available on the authors' 
website\footnote{\url{www-etud.iro.umontreal.ca/~boulanni/icml2012}}.

\subsubsection{Setup and Results}

\begin{table*}[t] 
\caption{Log-likelihood and expected accuracy for various RNN models in the symbolic music prediction task.
The double line separates sigmoid recognition layers (above) to structured output probability models (below).}
\label{tab:mus-res}
\vskip 0.1in
\centering
\begin{tabular}{lrrrrrrrrrrrr}
\hline \noalign{\smallskip}
Model & \multicolumn{3}{c}{Piano-midi.de} & \multicolumn{3}{c}{Nottingham} & \multicolumn{3}{c}{MuseData} & \multicolumn{3}{c}{JSB chorales} \\
 & \multicolumn{1}{c}{LL} & \multicolumn{1}{c}{LL} & \multicolumn{1}{c}{ACC \%} 
 & \multicolumn{1}{c}{LL} & \multicolumn{1}{c}{LL} & \multicolumn{1}{c}{ACC \%} 
 & \multicolumn{1}{c}{LL} & \multicolumn{1}{c}{LL} & \multicolumn{1}{c}{ACC \%} 
 & \multicolumn{1}{c}{LL} & \multicolumn{1}{c}{LL} & \multicolumn{1}{c}{ACC \%} \\ 
 & (train) & (test) & (test) 
 & (train) & (test) & (test) 
 & (train) & (test) & (test) 
 & (train) & (test) & (test)\\ 
\hline \noalign{\smallskip}

RNN (SGD)        &     -7.10 &     -7.86 &     22.84 &     -3.49 &     -3.75 &     66.90 &     -6.93 &     -7.20 &     27.97 &    -7.88&    -8.65 & \bf 29.97 \\
RNN (SGD+C)      &     -7.15 & \bf -7.59 &     22.98 &     -3.40 &     -3.67 &     67.47 &     -6.79 &     -7.04 &     30.53 &    -7.81&    -8.65 & \bf 29.98 \\
RNN (SGD+CL)     &     -7.04 & \bf -7.57 &     22.97 &     -3.31 &     -3.57 &     67.97 & \bf -6.47 & \bf -6.99 & \bf 31.53 &    -7.78&    -8.63 & \bf 29.98 \\
RNN (SGD+CLR)    & \bf -6.40 &     -7.80 & \bf 24.22 & \bf -2.99 &     -3.55 & \bf 70.20 &     -6.70 &     -7.34 &     29.06 &\bf -7.67&    -9.47 & \bf 29.98 \\
RNN (SGD+CRM)    &     -6.92 &     -7.73 &     23.71 &     -3.20 & \bf -3.43 &     68.47 &     -7.01 &     -7.24 &     29.13 &  -8.08  &    -8.81 &     29.52 \\
RNN (HF)         &     -7.00 & \bf -7.58 &     22.93 &     -3.47 &     -3.76 &     66.71 &     -6.76 &     -7.12 &     29.77 &  -8.11  & \bf-8.58 &     29.41 \\
\hline \hline \noalign{\smallskip}
RNN-RBM            &     N/A &     -7.09 & \bf 28.92 &    N/A  &    -2.39 & \bf 75.40 &   N/A   &    -6.01 & \bf 34.02  &     N/A  &    -6.27 &    33.12  \\
RNN-NADE (SGD)     &   -7.23 &     -7.48 &     20.69 &   -2.85 &    -2.91 &     64.95 &   -6.86 &    -6.74 &     24.91  &    -5.46 &    -5.83 &    32.11 \\
RNN-NADE (SGD+CR)  &   -6.70 &     -7.34 &     21.22 &   -2.14 &    -2.51 &     69.80 &   -6.27 &    -6.37 &     26.60  &    -4.44 &    -5.33 &    34.52 \\
RNN-NADE (SGD+CRM) &   -6.61 &     -7.34 &     22.12 &   -2.11 &    -2.49 &     69.54 &   -5.99 &    -6.19 &     29.62  &\bf -4.26 &\bf -5.19 &\bf 35.08 \\
RNN-NADE (HF)      &\bf -6.32& \bf -7.05 &     23.42 &\bf -1.81&\bf -2.31 &     71.50 &\bf -5.20&\bf -5.60 &     32.60  &    -4.91 &    -5.56 &    32.50 \\

\end{tabular}
\end{table*}

\begin{table*}[t] 
\caption{Entropy (bits per character) and perplexity for various RNN models 
on next character and next word prediction task. }
\label{tab:text-res}
\vskip 0.1in
\begin{center}
\begin{tabular}{lcccc}
\hline \noalign{\smallskip}
Model & \multicolumn{2}{c}{Penn Treebank Corpus} & \multicolumn{2}{c}{Penn Treebank Corpus} \\
& \multicolumn{2}{c}{word level} & \multicolumn{2}{c}{character level} \\
 & \multicolumn{1}{c}{perplexity} & \multicolumn{1}{c}{perplexity} & \multicolumn{1}{c}{entropy} & \multicolumn{1}{c}{entropy}  \\
 & \multicolumn{1}{c}{(train)} & \multicolumn{1}{c}{(test)} & \multicolumn{1}{c}{(train)} & \multicolumn{1}{c}{(test)}  \\
\hline \noalign{\smallskip}

RNN (SGD)        &   112.11  &   145.16  &    1.78   &   1.76         \\
RNN (SGD+C)      &    78.71  &   136.63  & \bf   1.40   & \bf  1.44         \\
RNN (SGD+CL)     &    76.70  &   129.83  &    1.56   &   1.56         \\
RNN (SGD+CLR)    & \bf 75.45  & \bf  128.35  &    1.45   &   1.49         \\
\end{tabular}
\end{center}
\end{table*}

We select hyperparameters, such as the number of hidden units $n_h$, regularization coefficients $\lambda_{L1}$, the choice of non-linearity function, or the momentum schedule $\mu_t$, 
learning rate $\epsilon_t$, number of leaky units $n_{leaky}$ or leaky factors $\alpha$
according to log-likelihood on a validation set and we report the final performance on the test set for the best choice in each category.
We do so by using random search~\cite{Bergstra+Bengio-2012-small}  on the following intervals:
\begin{tabbing}
$n_h\in [100, 400]$ \hspace{1.8cm}\= $\epsilon_t\in[10^{-4}, 10^{-1}]$  \\
$\mu_t \in[10^{-3}, 0.95]$ \>  $\lambda_{L1} \in[10^{-6}, 10^{-3}]$   \\
$n_{leaky}\in \{0\%, 25\%, 50\%\}$ \> $\alpha\in [0.02, 2]$
\end{tabbing}

The cutoff threshold for gradient clipping is set based on the average norm of the gradient over one pass on the data, and we used 
15 in this case for all music datasets. The data is split into sequences of 100 steps over which we compute the gradient. The hidden 
state is carried over from one sequence to another if they belong to the same song, otherwise is set to 0.

Table~\ref{tab:mus-res} presents log-likelihood (LL) and expected frame-level accuracy for various RNNs in the symbolic music prediction task. 

Results clearly show that these enhancements allow to improve on regular SGD in almost all cases; they also make SGD competitive with HF for the sigmoid recognition layers RNNs.

\subsection{Text Data}

We use the Penn Treebank Corpus to explore both word and character prediction tasks. 
The data is split by using sections 0-20 as training data (5017k characters), 
sections 21-22 as validation (393k characters) and sections 23-24 as test data (442k 
characters). 

For the word level prediction, we fix the dictionary to 10000 words, which we divide into 
30 classes according to their frequency in text (each class holding approximately 3.3\% of the total 
number of tokens in the training set). Such a 
factorization allows for faster implementation, as we are not required to 
evaluate the whole output layer (10000 units) which is the computational bottleneck, but only 
the output of the corresponding class~\cite{Mikolov-ICASSP-2011}.

\subsubsection{Setup and Results}

In the case of next word prediction, we compute gradients over sequences of 40 steps, where we carry the hidden 
state from one sequence to another. 
We use a small grid-search around the parameters used to get state of the art results for this number of classes 
\cite{Mikolov-ICASSP-2011}, i.e., with a network of 200 hidden units yielding 
a perplexity of 134. We explore learning rate
of $0.1, 0.01, 0.001$, rectifier units versus sigmoid units, cutoff threshold for the gradients of 30, 50 or none,
and no leaky units versus 50 of the units being sampled from 0.2 and 0.02. 

For the character level model we compute gradients over sequences of 150 steps, as we assume that longer dependencies are 
more crucial in this case. We use 500 hidden units and explore learning rates of 0.5, 0.1 and 0.01. 

In table~\ref{tab:text-res} we have entropy (bits per character) or  perplexity for varous RNNs on 
the word and character prediction tasks.  Again, we observe substantial improvements
in both training and test perplexity, suggesting that these techniques make
optimization easier.

\section{Conclusions}

Through our experiments we provide evidence that part of the issue of
training RNN is due to the rough error surface which can not be easily
handled by SGD. We follow an incremental set of improvements to SGD, and
show that in most cases they improve both the training and test error,
and allow this enhanced SGD to compete or even improve on a second-order
method which was found to work particularly well for RNNs, i.e., Hessian-Free
optimization.

\bibliographystyle{IEEEbib}
\bibliography{strings,ml,aigaion}

\end{document}